\documentclass{article} % For LaTeX2e
\usepackage{iclr2018_conference,times}
\usepackage{hyperref}
\usepackage{url}
\usepackage{polski}
\usepackage[utf8]{inputenc}

\title{Novel Ranking-Based Lexical Similarity Measure for Word Embedding}

\author{Jakub Dutkiewicz \& Czesław Jędrzejek \\
	Politechnika Poznańska\\
	Poznań, Poland \\
	\texttt{\{jakub.dutkiewicz,czeslaw.jedrzejek\}@put.poznan.pl}
}

\begin{document}

	\maketitle
	
	\begin{abstract}
	Distributional semantics models derive word space from linguistic items in context. Meaning is obtained by defining a distance  measure between vectors corresponding to lexical entities. Such vectors present several problems. In this paper we provide a guideline for post process improvements to the baseline vectors. We focus on refining the similarity aspect, address imperfections of the model by applying the hubness reduction method, implementing relational knowledge into the model, and providing a new ranking similarity definition  that give maximum weight to the top 1 component value. This feature ranking  is similar to the one used in   information retrieval.   All these enrichments outperform current literature results for joint ESL and TOEF sets comparison. Since single word embedding is a basic element of any semantic task one can expect a significant improvement of results for these tasks. Moreover, our improved  method of text processing can be translated to continuous distributed representation of biological sequences for deep proteomics and genomics.
	\end{abstract}
	
	\section{Introduction}
	
	Distributional language models are frequently used to measure word similarity in natural language (e.g. \cite{Frackowiak2017}). Recent works usually use the Distributional Hypothesis (\cite{harris54}) to generate the language models. This model often consists of a set of vectors; each vector corresponds to a character string, which represents a word. \cite{DBLP:journals/corr/abs-1301-3781} and \cite{pennington-etal-2014} implement word embedding (WE) algorithms. Vector components in language models created by these algorithms are latent. Similarity between words is defined as a function of vectors corresponding to the given words. The cosine measure is the most frequently used similarity function. \cite{DBLP:journals/corr/SantusCLLH16} highlights the fact that the cosine can be outperformed by ranking based functions. Vector space word representations obained from purely distributional information of words in large unlabelled corpora are not enough to best the state-of-the-art results in query answering benchmarks, because they suffer from 4 types of weaknesses:
	\begin{enumerate}
		\item Inadequate definition of similarity,
		\item Inability of accounting of senses of words,
		\item Appeareance of hubness that distorts distances between vectors,
		\item Inability of distinguishing from antonyms
	\end{enumerate}

	In this paper we use the existing word embedding model but with several post process enhancement techniques. We address three out of four of these issues. In particular we define novel similarity measure, dedicated for language models. The Euclidean distance is based on the locations of points in such a space.
	
	Similarity is a function, which is an monotonically opposite to distance. As the distance between two given entities gets shorter, entities are more similar. This holds for language models. Similarity between words is equal to similarity between their corresponding vectors. There are various definitions of distance. The most common Euclidean distance is defined as follows:
	\begin{equation}
	d(p_1,p_2) = \sqrt{\sum_{c \in p}(c_{p_1}-c_{p_2})^2}
	\end{equation}
	Similarity based on the Euclidean definition is inverse to the distance:
	\begin{equation}
	sim(p_1,p_2) = \frac{1}{1+d(p_1,p_2)}
	\end{equation}
	Angular definition of distance is defined with cosine function:
	\begin{equation}
	d(p_1,p_2) = 1-cos(p_1,p_2)
	\end{equation}
	We define angular similarity as:
	\begin{equation}
	sim(p_1,p_2) = cos(p_1,p_2)
	\end{equation}
	Both Euclidean and Cosine definitions of distance could be looked at as the analysis of vector components. Simple operations, like addition and multiplication work really well in low dimensional spaces. We believe, that applying those metrics in spaces of higher order is not ideal, hence we compare cosine similarity to a measure of distance dedicated for high dimensional spaces.

	\section{Related work}
	
	\cite{DBLP:journals/corr/SantusCLLH16} introduce the ranking based similarity function called APSyn. In their experiment APSyn outperforms the cosine similarity, reaching 73\% of accuracy in the best cases (an improvement of 27\% over cosine) on the ESL dataset, and 70\% accuracy (an improvement of 10\% over cosine) on the TOEFL dataset. In contrast to our work, they use the Positive Pointwise Mutual Information algorithm to create their language model.
	
	A successful avenue to enhance WE was pointed out by \cite{DBLP:journals/corr/FaruquiDJDHS14}, using WordNet (\cite{Miller2007}), and the Paraphrase Database (\cite{Ganitkevitch13ppdb}) to provide synonymy relation information to vector optimization equations. They call this process retrofitting, a pattern we adapt to the angular definition of distance, which is more suitable to our case.
	
	We also address hubness reduction. Hubness is related to the phenomenon of concentration of distances - the fact that points get closer at large vector dimensionalities. Hubness is very pronounced for vector dimensions of the order of thousands. We apply this method of localized centering for hubness reduction (\cite{Feldbauer2016}) for the language models.
	\section{Method}
	
	In our work we define the language model as a set of word representations. Each word is represented by its vector. We refer to a vector corresponding to a word $w_i$ as $v_i$. A complete set of words for a given language is referred to as a vector space model. We define similarity between words $w_i$ and $w_j$ as a function of vectors $v_i$ and $v_j$.
	
	\begin{equation}
	sim(w_i,w_j) = f(v_i,v_j)
	\end{equation}
	We present an algorithm for obtaining optimized similarity measures given a vector space model for word embedding. The algorithm consists of 6 steps:
	\begin{enumerate}
		\item{Refine the vector space using the L2 retrofit algorithm}
		\item{Obtain vector space of centroids}
		\item{Obtain vectors for a given pair of words and optionally for given context words}
		\item{Recalculate the vectors using the localized centering method}
		\item{Calculate ranking of vector components for a given pair of words}
		\item{Use the ranking based similarity function to obtain the similarity between a given pair of words.}
	\end{enumerate}
	
	We use all of the methods in together to achieve significant improvement over the baseline method. We present details of the algorithm in the following sections.
	\subsection{Baseline}
	The cosine function provides the baseline similarity measure:
	
	\begin{equation}
	sim(w_1,w_2) = cos(v_1,v_2) = \frac{v_1 \cdot v_2}{\|v_1\| \|v_2\|}
	\end{equation}
	The cosine function has been  to achieving a reasonable baseline. It is superior to the Euclidean similarity measure and is used in various works related to word similarity. In our work we use several post-process modifications to the vector space model; we also redefine the similarity measure. 
	
	\subsection{Implementing relational knowledge into the vector space model}
	Let us define a lexicon of synonyms $L$. Each row in the lexicon consists of a word and a set of its synonyms.
	
	\begin{equation}
	L(w_i)=\{w_j \colon synonymity(w_i,w_j) \}
	\end{equation}

	A basic method of implementing synonym knowledge into the vector space model was previously described in \cite{Ganitkevitch13ppdb}. We refer to that method as retrofit; it uses the iterational algorithm of moving the vector towards an average vector of its synonyms according to the following formula.
	\begin{equation}
	v_i' = \frac{\alpha_iv_i + \frac{\sum_{w_j \in L_(wj)} \beta_jv_j}{\|L(w_j)\|}}{2}
	\end{equation}
	In the original formula \cite{DBLP:journals/corr/FaruquiDJDHS14}, variables $\alpha$ and $\beta$ allow us to weigh the importance of certain synonyms. The basic retrofit method moves the vector towards its destination (shortens the distance between the average synonym to a given vector) using the Euclidean definition of distance. This is not consistent with the cosine distance. Instead we improve \cite{DBLP:journals/corr/FaruquiDJDHS14} idea by performing operations in spherical space by normalizing the vector, thus preserving the angular definition of distance. This amounts to rotating the vector instead of translating it. We implemented the basic transformation for the rotation; however it proved to be time consuming, which affected our work on the subject. Therefor, for simplicity, the average vector of two normalized vectors is precisely between given vectors in both the Euclidean and angular definition of distance. This gives the following formula:
	\begin{equation}
	v_i' = \|v_i\|\frac{\frac{v_i}{\|v_i\|} + \frac{\sum_{w_j \in L(w_j)} \frac{v_j}{\|v_j\|}}{\|L(w_j)\|}}{2}
	\end{equation}
	We refer to this method as to L2 retrofitting.

	\subsection{Localized centering}
	
	We address the problem of hubness in high dimensional spaces with the localized centering approach applied to every vector in the space. The centered values of vectors, centroids, are the average vectors of $k$ nearest neighbors of the given vector $v_i$. We apply a cosine distance measure to calculate the nearest neighbors.
	\begin{equation}
	c_i=\frac{\sum_{v_j \in k-NN(v_i)}v_i )}{N}
	\end{equation}
	In \cite{Feldbauer2016}, the authors pointed out that skewness of a space has a direct connection to the hubness of vectors. We follow the pattern presented in that work and recalculate the vectors using the following formula.
	
	\begin{equation}
	v_i' = v_i - c_i^\gamma
	\end{equation}
	
	Parameter $\gamma$ in the equation is equal to the skewness of the space.  
	
	\subsubsection{Ranking based similarity function}
	We propose a component ranking function as the similarity measure. This idea was originally introduced in \cite{DBLP:journals/corr/SantusCLLH16} who proposed the APSyn ranking function. Let us define the vector $v_i$ as a list of its components.
	\begin{equation}
	v_i=[f_1,f_2,...,f_n ]
	\end{equation}
	We then obtain the ranking $r_i$ by sorting the list in descending order (d in the equation denotes type of ordering), denoting each of the components with its rank on the list.
	\begin{equation}
	r_i^d = \{f_1:rank_i^d(f_1), ... ,f_n:rank_i^d(f_n)\}
	\end{equation}
	
	APSyn is calculated on the intersection of the N components with the highest score.
	\begin{equation}
	APSyn(w_i,w_j) = \sum_{f_k \in top(r_i^d) \cap top(r_j^d)}\frac{2}{rank_i(f_k)+rank_j(f_k)}
	\end{equation}
	
	APSyn was originally computed on the PPMI language model, which has unique feature of non-negative vector components. As this feature is not given for every language model, we take into account negative values of the components. We define the negative ranking by sorting the components in ascending order (a in the equation denotes type of ordering).
	
	\begin{equation}
	r_i^a = \{f_1:rank_i^a(f_1), ... ,f_n:rank_i^a(f_n)\}
	\end{equation}
	
	As we want our ranking based similarity function to preserve some of the cosine properties, we define score values for each of the components and similarly to the cosine function, multiply the scores for each component. As the distribution of component values is Gaussian, we use the exponential function.

	\begin{equation}
	s_{i,f_k}=e^{-rank_i (f_k)\frac{k}{d}}
	\end{equation}
	
	Parameters $k$ and $d$ correspond respectively to weighting of the score function and the dimensionality of the space. With high k values, the highest ranked component will be the most influential one. The rationale is maximizing information gain. Our measure is similar to infAP and infNDCG measures used in information retrieval \cite{doi:10.1093/database/bax068}  that give maximum weight to the top 1 result. P@10 gives equal weight to the top 10,results.  Lower $k$ values increase the impact of lower ranked components  at the expense of ‘long tail’ of ranked components. We use the default $k$ value of 10. The score function is identical for both ascending and descending rankings. 
	We address the problem of polysemy with a differential analysis process. Similarity between pair of words is captured by discovering the sense of each word and then comparing two given senses of words. The sense of words is discovered by analysis of their contexts. We define the differential analysis of a component as the sum of all scores for that exact component in each of the context vectors.
	
	\begin{equation}
	h_{i,f_k} = \sum_{w_j \in context(w_j)}s_{j,f_k}
	\end{equation}
	
	Finally we define the Ranking based Exponential Similarity Measure (RESM) as follows.
	\begin{equation}
	RESM^a(w_i,w_j) = \sum_{f_k \in top(r_i^d) \cap top(r_j^d)}\frac{s_{i,f_k}^a s_{j,f_k}^a}{h_{i,f_k}^a}
	\end{equation}
	The equation is similar to the cosine function. Both cosine and RESM measures multiply values of each component and sum the obtained results. Contrary to the cosine function, RESM scales with a given context. It should be noted, that we apply differential analysis with a context function $h$. An equation in this form is dedicated for the test sets we use in evaluation. The final value is calculated as a sum of the RESM for both types of ordering.
	
	\begin{equation}
	RESM(w_i,w_j) = RESM^a(w_i,w_j) + RESM^d(w_i,w_j)
	\end{equation}
	
	\subsection{Implementation}
	
	The algorithm has been implemented in C\#. It is publicly available via the repository along with implementation of the evaluation.\footnote{https://github.com/dudenzz/DistributionalModel} 
	
	\section{Evaluation}
	\begin{table}
		\caption{Example questions.}
		\label{questions}
		\begin{center}
			\begin{tabular} {|c|c|c|c|c|}
				\hline
				Q.word & P1 & P2 & P3 & P4  \\ \hline 
				Iron & Wood & Metal & Plastic &  Stone \\ \hline 
				Iron & Wood & Crop & Grass & Arrow \\ \hline 
			\end{tabular}
		\end{center}
	\end{table}
	\begin{table}
		\caption{State of the art results for TOEFL and ESL test sets}
		\label{sota}
		\begin{center}
			\begin{tabular} {|c|c|c|}
				\hline
				\cite{Bullinaria2012}  & 100.0\% & 66.0\% \\
				\cite{sterlund2015FactorizationOL} & & \\\hline 
				\cite{DBLP:journals/corr/abs-1204-0245} & 79.7\% & 82.0\% \\ \hline 
				\cite{Lu2011UsingFS} & 97.5\% & 86.0\% \\ \hline 
			\end{tabular}
		\end{center}
	\end{table}
	We have tested our method against TOEFL and ESL test sets. TOEFL consists of 80 questions, ESL consists of 50 questions. Questions in ESL are significantly harder. Both tests consist of questions designed for nonnative speakers of English. Each question in the tests consists of a question word with  a set of four answers. It is worth pointing out, that the context given by a set of possible answers often defines the question. Example questions in Table \ref{questions} highlight the problem. In the first question, all of possible answers are building materials. Wood should be rejected as there is more appopriate answer. In second question, out of possible answers, only wood is a building material which makes it a good candidate for the correct answer. This is a basis for applying a differential analysis in the similarity measure. Table \ref{sota}. illustrates state of the art results for both test sets. The TOEFL test set was introduced in \cite{Landauer97asolution}; the  ESL test set was introduced in \cite{Turney2001}
	
	\subsection{Experimental setup}
	
	We use the unmodified vector space model trained on 840 billion words from Common Crawl data with the GloVe algorithm introduced in \cite{pennington-etal-2014}. The model consists of 2.2 million unique vectors; Each vector consists of 300 components. The model can be obtained via the GloVe authors website.
	We run several experiments, for which settings are as follows: In the evaluation skewness $\gamma = 9$ and $k=10$. All of the possible answers are taken as context words for the differential analysis. In our runs we use all of the described methods separately and conjunctively. We refer to the methods in the following way. We denote the localized centering method for hubness reduction as HR. We use a Paraphrase Database lexicon introduced in \cite{Ganitkevitch13ppdb} for the retrofitting. We denote L2 retrofitting as RETRO.
	
	\subsection{Heuristic improvement}
	Although the hubness reduction method does not increase the number of correct answers for the ESL test set, we noticed that the average rank of the correct answer goes down from 1.32 to 1.24. That is a significant improvement. To obtain better results we combined the results with and without localized centering. The heuristical method chooses the answer with the best average rank for both sets. By applying that method we obtained two additional correct answers. 
	\begin{table}
		\caption{Accuracy of various methods on TOEFL and ESL test sets }
		\label{ourResults}
		\begin{center}
			\begin{tabular} {|c|c|c|}
				\hline
				Method & TOEFL & ESL \\ \hline
				Cosine  & 88.75\% & 60.00\% \\\hline
				HR + Cosine & 91.25\% & 66.00\% \\\hline 
				RETRO + Cosine & 95.00\% & 62.00\% \\ \hline 
				HR + RETRO + Cosine  & 96.25\% & 74.00\% \\ \hline
				APSyn & 80.00\%	& 60.00\% \\ \hline
				RETRO + APSyn & 97.50\% & 70\% \\ \hline 
				RESM & 90.00\% & 76.00\% \\ \hline
				RETRO + RESM & 96.25\% & 80.00\% \\ \hline
				HR + RETRO + RESM & 97.50\% & 80.00\% \\ \hline
				RETRO + RESM + heuristic& 97.50\% & 84.00\% \\ \hline 
			\end{tabular}
		\end{center}
	\end{table}
	
	We improved the accuracy results by 8.75\% and 24\% for TOEFL and ESL test sets respectively. We observe the largest improvement of accuracy by applying the localized centering method for TOEFL test set. Testing on the ESL question set seems to give the best results by changing the similarity measure from cosine to RESM. Thus  each step of the algorithm improves the results. The significance of the improvement varies based  on the test set. A complete set of measured accuracies is presented in Table \ref{ourResults}. The results for the APSyn ranking method are obtained using the Glove vector, not using PPMI as in /cite{DBLP:journals/corr/SantusCLLH16}
	
	\section{Conclusions}
	
    The procedure we provide in this  is capable of achieving the best results in the word embedding category and (nearly) state-of-the art results for any current metod. The work of \cite{Lu2011UsingFS} employed 2 fitting constants (and it is not clear that they were the same for all questions) for answering the TOEFL test where only 50 questions are used. Techniques introduced in the paper are lightweight and easy to implement, yet they provide a significant performance boost to the language model. Recently, a lot of progress was achieved in relating antonyms to synonyms \cite{DBLP:journals/corr/SantusCLLH16}, \cite{nguyen}. We tried the antonym RETRO method to take into account relational knowledge on antonyms. The method repels two vectors that are an antonym pair. Contrary to \cite{nguyen}, who obtained minimal improvement, in our case  accuracy does not improve (for any value of repulsion strength), which could be because of relatively large window (10) in the original Glove work. All works that could distingush antonyms from synonyms using word embedding used much smaller context windows. We plan to extent our Glove based calculations to smaller windows. This needs to be studied more thoroughly.
    Since the single word embedding is a basic element of any semantic task one can expect a signicant improvement of results for these tasks.  In particular,  SemEval-2017 
    International Workshop on Semantic Evaluation run (among others) the following tasks(\cite{se2017}):
    \begin{enumerate}
    	\item Task 1: Semantic Textual Similarity
    	\item Task 2: Multilingual and Cross-lingual Semantic Word Similarity
    	\item Task 3: Community Question Answering
    \end{enumerate}
     
in the category Semantic comparison for words and texts.
Another immediate application would be information retrieval (IR).
Expanding queries by adding potentially relevant terms is a common practice in improving relevance in IR systems. There are many methods of query expansion. Relevance feedback takes the documents on top of a ranking list and adds terms appearing in these document to a new query. In this work we use the idea to add synonyms and other similar terms to query terms before the pseudo- relevance feedback. This type of expansion can be divided into two categories. The first category involves the use of ontologies or lexicons (relational knowledge). The second category is word embedding (WE). Here closed words  for expansion have to be very precise, otherwise a query drift may occur, and precision and accuracy of retrieval may deteriorate.

Moreover, our improved  method of text processing can be translated to continuous distributed representation of biological sequences for deep proteomics and genomics. Protein sequence is typically notated as a string of letters, listing the amino acids starting at the amino-terminal end through to the carboxyl-terminal end. Either a three letter code or single letter code can be used to represent the 20 naturally occurring amino acids. Until recently most methods used n-grams. The immediate application is family classification task \cite{10.1371/journal.pone.0141287}.

	\subsubsection*{Acknowledgement}
This work was supported by the PUT DS grant no 04/45/DSPB/0149
	
	\bibliography{resm_bib}
	\bibliographystyle{iclr2018_conference}
	
\end{document}